\documentclass[11pt,a4paper]{article}
\usepackage[hyperref]{acl2019}
\usepackage{times}
\usepackage{latexsym}
\usepackage{xcolor}
\usepackage{amsmath}
\usepackage{tabularx}
\usepackage{float}
\usepackage{graphicx}
\usepackage{natbib}

\usepackage{url}
\usepackage{hyperref}

\aclfinalcopy 

\newcommand{\ignore}[1]{}

\newcommand{\dr}[1]{\textcolor{red}{[#1---dr]}}

\title{Improving Generalization in Coreference Resolution via Adversarial Training}
\author{Sanjay Subramanian \\
  University of Pennsylvania \\
  \texttt{sanjayssub34@gmail.com} \\\And
  Dan Roth \\
  University of Pennsylvania \\
  \texttt{danroth@seas.upenn.edu}}

\date{}

\begin{document}
\maketitle
\begin{abstract}
  \ignore{\dr{I like the first presentation better, since it flows better; I don't feel strongly about it, so if you like the revised version better, it's ok. Either way, I like the addition of the last sentence in the new version so include it. Also, you don't need to include the references in the abstract; you get to it anyhow in the paper; easier to read this way. (this means also changing "the SOTA of" to "a SOTA system"}}
  
  In order for coreference resolution systems to be useful in practice, they must be able to generalize to new text. In this work, we demonstrate that the performance of the state-of-the-art system decreases when the names of PER and GPE named entities in the CoNLL dataset are changed to names that do not occur in the training set. We use the technique of adversarial gradient-based training to retrain the state-of-the-art system and demonstrate that the retrained system achieves higher performance on the CoNLL dataset (both with and without the change of named entities) and the GAP dataset.\\
  \ignore{Revised version: In order for coreference resolution systems to be useful in practice, they must be able to generalize to new text. In this work, we show that adversarial gradient-based training \cite{Miyato16} provides a moderate improvement in the generalization of the state-of-the-art system by \cite{Lee2018}. We also introduce a new method of evaluating generalization that analyzes the degree to which the system is robust to names not seen in training. We show that the performance of the \cite{Lee2018} system decreases when the names of PER and GPE named entities in the CoNLL-2012 dataset \cite{Pradhan2012} are changed to names that do not occur in the training set. The system retrained with the adversarial gradient method achieves state-of-the-art performance on the CoNLL dataset (both with and without the change of named entities) and the GAP dataset \citep{gapp}.}
 \end{abstract}

%

\section{Introduction}
Through the use of neural networks, performance on the task of coreference resolution has increased significantly over the last few years.
%
Still, neural systems trained on the standard coreference dataset have issues with generalization, as shown by \cite{Moosavi2018b}.\\
One way to improve the understanding of how a system overfits a dataset is to study the change in the system's performance when the dataset is modified slightly in a focused and relevant manner. We take this approach by modifying the test set so that each PER and GPE (person and geopolitical entity) named entity is different from those seen in training. In other words, we ensure that there is no leakage of PER and GPE named entities from the training set into the test set. We demonstrate that the performance of the \cite{Lee2018} system, which is the current state-of-the-art, decreases when the named entities are replaced. An example of a replacement that causes the system to make an error is given in Table \ref{tab:example}.
\begin{table}
	\centering
	\small
	\begin{tabular}{|p{70mm}|}
			\hline
			\textbf{Original}: But \textcolor{blue}{\textbf{Dirk Van Dongen , president of the National Association of Wholesaler - Distributors}} , said that last month 's rise `` is n't as bad an omen '' as the 0.9 \% figure suggests . `` If you examine the data carefully , the increase is concentrated in energy and motor vehicle prices , rather than being a broad - based advance in the prices of consumer and industrial goods , '' \textcolor{blue}{\textbf{he}} explained . \\
			\hline
			\textbf{Replacement}: Replace \textit{Dirk Van Dongen} with \textit{Vendemiaire Van Korewdit}. \\
			\hline
	\end{tabular}
	\caption{An excerpt from the CoNLL test set. The coreference between the two highlighted mentions is correctly predicted by the \cite{Lee2018} system, but after the specified replacement, the system incorrectly resolves ``he'' to a different name occurring outside this excerpt.}
	\label{tab:example}
\end{table}
\ignore{\dr{I think the "One way to improve the understanding..." paragraph should be here. Then you can say, "motivated by these observations this papers attempts to improve the training process of coref systems." and continue as below.}
\ignore{the dataset cannot be used in isolation for the evaluation of coreference systems}}

\noindent
Motivated by these issues of generalization, this paper aims to improve the training process of neural coreference systems. Various regularization techniques have been proposed for improving the generalization capability of neural networks, including dropout \citep{Dropout} and adversarial training \citep{Goodfellow14, Miyato16}. The model of \cite{Lee2018}, like most neural approaches, uses dropout. In this work, we apply the adversarial fast-gradient-sign-method (FGSM) described by \cite{Miyato16} to the model of \cite{Lee2018}, and show that this technique improves the model's generalization even when applied on top of dropout.\\
\noindent The CoNLL-2012 Shared Task dataset \citep{Pradhan2012} has been the standard dataset used for both training and evaluating English coreference systems since the dataset was introduced. The dataset includes seven genres that span multiple writing styles and multiple nationalities. We demonstrate that the system of \cite{Lee2018} retrained with adversarial training achieves state-of-the-art performance on the original CoNLL-2012 dataset \citep{Pradhan2012} as well as the CoNLL-2012 dataset with changed named entities. Furthermore, the system trained with the adversarial method exhibits state-of-the-art performance on the GAP dataset \citep{gapp}, a recently released dataset focusing on resolving pronouns to people's names in excerpts from Wikipedia. The code and other relevant files for this project can be found via \href{https://cogcomp.org/page/publication_view/871}{https://cogcomp.org/page/publication\_view/871}.

\section{Related Work}
\cite{Moosavi2017a,Moosavi2018b} also study generalization of neural coreference resolvers. However, they focus on transfer and indicate that the ranking of coreference resolvers (trained on the CoNLL training set) induced by their performance on the CoNLL test set is not preserved when the systems are evaluated on a different dataset. They use the Wikicoref dataset \citep{Wikicoref}, which is limited in that it consists of only $ 30 $ documents. 
They then show that the addition of features representing linguistic information improves the performance of a coreference resolver on the out-of-domain dataset. 

\noindent
The adversarial fast-gradient-sign-method (FGSM) was first introduced by \cite{Goodfellow14} and was applied to sentence classification tasks through word embeddings by \cite{Miyato16}. Gradient-based adversarial attacks have since been used to train models for various NLP tasks, such as relation extraction \citep{Wu17} and joint entity and relation extraction \cite{Bekoulis18}. 

Our replacements of named entities can also be viewed as a way of generating adversarial examples for coreference systems;
it is related to the earlier method proposed in \cite{KKSCER16} in the context of question answering and to \cite{AdvEx}, which provides a way of generating adversarial examples for simple classification tasks.

\section{Adversarial Training for Coreference}
In coreference resolution, the goal is to find and cluster phrases that refer to entities. We use the word ``span'' to mean a series of consecutive words. A span that refers to an entity is called a mention. If two mentions $ i $ and $ j $ refer to the same entity and mention $ i $ occurs before mention $ j $ in the text, we say that mention $ i $ is an antecedent of mention $ j $. For a given mention $ i $, the candidate antecedents of $ i $ are the mentions that occur before $ i $ in the text.
\begin{figure}
    \centering
    \includegraphics[width=60mm]{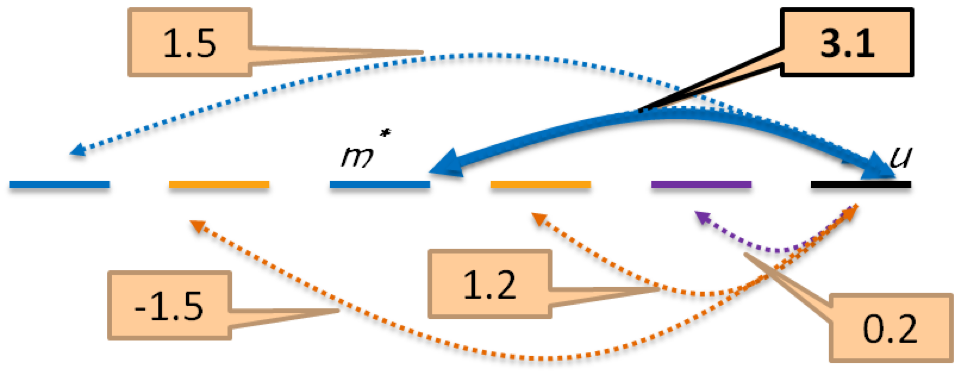}
    \caption{For each mention, the model computes scores for each of the candidate antecedent mentions and chooses the candidate with the highest score to be the predicted antecedent. This image was created by the authors of \citep{ChangSaRo13}.}
    \label{fig:coref_diagram}
\end{figure}
In Figure \ref{fig:coref_diagram}, each line segment represents a mention and the arrows are directed from one mention to its possible antecedents.\\
We now review the model architecture of \cite{Lee2018} and describe how we apply the fast-gradient-sign-method (FGSM) of \cite{Miyato16} to the model. Using GloVe \citep{glove} and ELMo \cite{elmo} embeddings of each word and using learned character embeddings, the model computes contextualized representations $ \{\mathbf{x}_1, \mathbf{x}_2, ..., \mathbf{x}_n\} $ of each word $x_i$ in the input document using a bidirectional LSTM \cite{lstm}. For candidate span $ i $, which consists of the words at indices $ start_i, start_i+1, ..., end_i $, the model constructs a span representation $ \mathbf{g}_i $ by concatenating $ \mathbf{x}_{start_i} $, $ \mathbf{x}_{end_i} $, $ \frac{1}{\sum_{j=start_i}^{end_i} \beta_j}\sum_{j=start_i}^{end_i} \beta_j\mathbf{x}_j $, and $ \phi(end_i-start_i) $, where the $ \beta_j $'s are learned scalar values and $ \phi(\cdot) $ is a learned embedding representing the width of the span \citep{Lee2017}. The span representations are then used as inputs to feedforward networks that compute mention scores for each span and that compute antecedent scores for pairs of spans. In Figure \ref{fig:coref_diagram}, the number associated with each arrow is the antecedent score for the associated pair of mentions. The coreference score for the pair of spans $ (i, j) $ is the sum of the mention score for span $ i $, the mention score for span $ j $, and the antecedent score for $ (i, j) $. For each span $ i $, the antecedent span predicted by the model is the span $ j $ that maximizes the antecedent score for $ (i, j) $. Let $ \mathbf{g} = \left\{\mathbf{g}_i\right\}_{i=1}^N $ denote the set of the representations of all $ N $ candidate spans. Let $ \mathcal{L}(\mathbf{g}) $ denote the original model's loss function. (Note that the model's predictions and the loss depend on the input text only through the span representations.) For each $ i \in \{1, ..., N\} $, let $ \mathbf{g}_i^{adv}(\mathbf{g}) = \nabla_{\mathbf{g}_i} \mathcal{L}\left(\left\{\mathbf{g}_i\right\}_{i=1}^N\right) $ denote the gradient of the loss with respect to the span embeddings. Then the adversarial loss with the FGSM is
\begin{align*}
	\mathcal{L}_{adv}(\{\mathbf{g}_i\}_{i=1}^N) =&\; \mathcal{L}\left(\left\{\mathbf{g}_i+\epsilon\frac{\mathbf{g}_i^{adv}(\mathbf{g})}{||\mathbf{g}_i^{adv}(\mathbf{g})||}\right\}_{i=1}^N\right).
\end{align*}
The total loss used in training is
\begin{align*}
	\mathcal{L}_{total}(\mathbf{g}) =&\; \alpha\mathcal{L}\left(\mathbf{g}\right)+(1-\alpha)\mathcal{L}_{adv}\left(\mathbf{g}\right).
\end{align*}
In our experiments, we find that $ \alpha = 0.6 $ and $ \epsilon = 1 $ work well. A key difference between our method and that employed by \cite{Miyato16} is that the latter applies the adversarial perturbation to the input embeddings, whereas we apply it to the span representations, which are an intermediate layer of the model. We found in our experiments that applying the FGSM to the  character embeddings in the initial layer was not as effective as applying the method to the span representations as described above. Another difference between our method and that of \cite{Miyato16} is that we do not normalize the span embeddings before applying the adversarial perturbations.

\section{No Leakage of Named Entities}
\label{sec:perturb}
Named entities are an important subset of the entities a coreference system is tasked with discovering. \cite{Agarwal18} provide the percentages of clusters in the CoNLL dataset represented by the PER, ORG, GPE, and DATE named entity types -- $ 15\% $, $ 11\% $, $ 11\% $, and $4\% $, respectively. It is important for generalization that systems perform well with names that are different from those seen in training. We found that in the CoNLL dataset, roughly $ 34\% $ of the PER and GPE named entities that are the head of a mention of some gold cluster in the test set are also the head of a mention of a gold cluster in the train set. Therefore, there is considerable overlap, or leakage, between the names in the train and test sets. In this section, we describe a method for evaluating on the CoNLL test set without leaked name entities.

\noindent
We focus on PER and GPE named entities because they are two of the three most common entity types and because in general when replacing a PER or GPE name with another name, it is easy to not change 
the true coreference structure of the document. In particular, changing the name of an organization while ensuring that it is compatible with nominals in the cluster is nontrivial without a finer semantic typing.
\begin{table*}[t!]
    \centering
    \begin{tabular}{|p{70mm}|p{70mm}|}
    \hline
    Original & No Leakage \\
    \hline
    We asked \textbf{Judy Muller} if she would like to do the story of a fascinating man . She took a deep breath and said , okay . & We asked \textbf{Sallie Kousonsavath} if she would like to do the story of a fascinating man . She took a deep breath and said , okay .\\
    \hline
    The last thing President \textbf{Clinton} did today before heading to the Mideast is go to church -- appropriate , perhaps , given the enormity of the task he and his national security team face in the days ahead . & The last thing President \textbf{Golia} did today before heading to the Mideast is go to church -- appropriate , perhaps , given the enormity of the task he and his national security team face in the days ahead .\\
    \hline
    In theory at least , tight supplies next spring could leave the wheat futures market susceptible to a supply - demand squeeze , said Daniel Basse , a futures analyst with AgResource Co. in \textbf{Chicago} . & In theory at least , tight supplies next spring could leave the wheat futures market susceptible to a supply - demand squeeze , said Daniel Basse , a futures analyst with AgResource Co. in \textbf{Machete} .\\
    \hline
    \end{tabular}
    \caption{Excerpts from the CoNLL-2012 test set and their versions after we have replaced PER and GPE names to avoid name leakage.}
    \label{tab:no_leakage_examples}
\end{table*}
By contrast, we describe below how we control for gender and location type when replacing PER and GPE names, respectively. We also ensure that the capitalization of the first letter in the replacement name is the same as in the original text. Finally, we note that the diversity of PER and GPE entities exceeds that of other named entity types; this increases the importance of generalization to new names and, at the same time, enables us to find matching names to use as replacements. Table \ref{tab:no_leakage_examples} provides examples of text in the original CoNLL-2012 dataset and the corresponding text after our modifications.
\ignore{By contrast, as we describe below, there exists a database that categorizes a large number GPE names by type of geographic entity (e.g. city, state). We ensure that the replacement for a GPE entity is of the same type so that nominals in the cluster will in general match the entity name. We do not replace country names in order to avoid inducing conflicts between the country name and associated adjective (e.g. ``Japan'' and ``Japanese''). With PER names, we describe below how we attempt to keep the gender of a name unchanged when replacing it, ensuring compatibility with gender-specific pronouns.}
\subsection{Replacing PER entities}
For replacing PER entities, we utilize the publicly available list of last names from the 1990 U.S. Census and a gazetteer of first names that has the proportion of people with this name who are males. The gazetteer was collected in an unsupervised fashion from Wikipedia. We denote the list of last names by $ \mathcal{L} $, the list of male first names (i.e. first names with male proportion greater than or equal to $ 0.5 $ in the gazetteer) by $ \mathcal{M} $, and the list of female first names (i.e. first names with male proportion less than or equal to $ 0.5 $ in the gazetteer) by $ \mathcal{F} $. We remove all names occurring in training from $ \mathcal{L} $, $ \mathcal{M} $, and $ \mathcal{F} $. We use the spaCy dependency parser \cite{spacy} to find the heads of each mention. We say that a mention is a person-mention if the head of the mention is a PER named entity, and we say that the name of the person-mention is the PER named entity that is its head. We use the dependency parser and the gold NER to identify all of the person-mentions. For each gold cluster containing a person-mention, we find the longest name among the names of all of the person-mentions in the cluster. If the longest name of a cluster has only one token, we assume that the name is a last name, and we replace the name with a name chosen uniformly at random from the remaining last names in $ \mathcal{L} $. Otherwise, if the longest name has multiple tokens, we say that the cluster is male if the cluster contains no female pronouns (``she'', ``her'', ``hers'') and one of the following is true: the first token does not appear in $ \mathcal{M} $ or $ \mathcal{F} $, if the token appears in $ \mathcal{M} $, or the cluster contains a male pronoun (``he'', ``him'', ``his''). We say that the cluster is female if it is not male. Then we (1) replace the last token with a name chosen uniformly at random from the remaining last names in $ \mathcal{L} $, and (2) replace the first token with a name chosen uniformly at random from the remaining first names in $ \mathcal{M} $ if the cluster is male or from the remaining first names $ \mathcal{F} $ if the cluster is female. Note that our sampling from each of $ \mathcal{L} $, $ \mathcal{M} $, and $ \mathcal{F} $ is without replacement, so no last name is used as a replacement more than once, no male first name is used more than once, and no female first name is used more than once.
\subsection{Replacing GPE entities}
Our approach to replacing GPE entity names is very similar to that used for PER names. We use the GeoNames\footnote{http://www.geonames.org/} database of geopolitical names. \ignore{\dr{Say how large it is so people will not be worried that your condition below is restricting.}} In addition to providing a list of GPE names, this database also categorizes the names by the type of entity to which they refer (e.g. city, state, county, etc.). The data includes the names and categories of more than $ 11,000,000 $ locations in the world. We restrict our attention to GPE entities that satisfy the following requirements: (1) they occur in the GeoNames database and (2) they are not countries. We say that a mention is a GPE-mention if its head (as given by the dependency parser) is a GPE named entity satisfying these three requirements. (Again, we use the gold NER to identify GPE names in the CoNLL text.) We remove all GPE names occurring in the training set from the list of replacement GPE names for each location category. Then for each cluster containing a GPE-mention, we find the GeoNames category for the mention's GPE name and replace the name with a randomly chosen name from the same category.  As with PER names, we sample names from each category without replacement, so each GPE name is used for replacement at most once.


\section{Experiments}
We trained the \cite{Lee2018} model architecture with the adversarial approach on the CoNLL training set for $ 355000 $ iterations (the same number of iterations for which the original model was trained) with the same training hyperparameters used by original model. For comparing with the \cite{Lee2017} and \cite{Lee2018} systems, we use the pretrained models released by the authors.\footnote{Available at \href{https://lil.cs.washington.edu/coref/final.tgz}{https://lil.cs.washington.edu/coref/final.tgz} and \href{http://lsz-gpu-01.cs.washington.edu/resources/coref/c2f_final.tgz}{https://lil.cs.washington.edu/coref/final.tgz}}

The datasets used for evaluation are the CoNLL and GAP datasets.

\subsection{CoNLL Dataset}
\begin{table}[t!]
	\begin{center}
		\begin{tabular}{|l|l|l|}
			\hline
			& \textbf{Original} & \textbf{No Leakage}  \\
			\hline
			\cite{Lee2018} & 72.96 & 71.84 \\
			\hline
			+Adv. Training & \textbf{73.23} & \underline{\textbf{72.32}} \\
			\hline
		\end{tabular}
	\end{center}
	\caption{Results (CoNLL F1) on the CoNLL Test Set. ``Original'' refers to the original test set, and ``No Leakage'' refers to the test set modified with the replacement of named entities described in Section \ref{sec:perturb}. For each dataset, highest score for each dataset is \textbf{bolded} and is \underline{underlined} if the difference between it and the other model's score is statistically significant ($ p < 0.21 $ per a stratified approximate randomization test similar to that of \cite{Noreen89}).}
	\label{tab:conllresults}
\end{table}

Table \ref{tab:conllresults} shows the performance on the CoNLL test set, as measured by CoNLL F1, of the \cite{Lee2018} system with and without our adversarial training approach.\footnote{Please note that the small differences between the No Leakage results here and those in the version of this paper in the ACL Anthology are due to a small mistake in our preprocessing pipeline, which we have fixed since publication.} The replacement of PER and GPE entities decreased the performance of the original system by more than $ 1 $ F1.

\begin{table}[t!]
	\begin{center}
		\begin{tabular}{|l|l|l|l|}
			\hline
			& \textbf{M} & \textbf{F} & \textbf{O} \\
			\hline
			\cite{Lee2017} & 68.7 & 60.0 & 64.5 \\
			\hline
			\cite{Lee2018} & 75.8 & 70.6 & 73.3 \\
			\hline
			+Adv. Training & \textbf{77.3} & \textbf{72.1} & \underline{\textbf{74.7}} \\
			\hline
		\end{tabular}
	\end{center}
	\caption{Results (F1 metric defined by \cite{gapp}) on the GAP Test Set. \textbf{M} refers to male pronouns, \textbf{F} refers to female pronouns, and \textbf{O} refers to the full evaluation data. For each category, highest score is \textbf{bolded} and underlined if difference between it and next-highest score is statistically significant ($ p < 0.05 $ per the McNemar test \cite{McNemarTest}).}
	\label{tab:gapresults}
\end{table}

\subsection{GAP Dataset}
The GAP dataset \citep{gapp} focuses on resolving pronouns to named people in excerpts from Wikipedia. The dataset, which is gender-balanced, consists of examples in which the system must determine whether a given pronoun refers to one, both, or neither of two given names. Thus, the task can be viewed a binary classification task in which the input is a (pronoun, name) pair and the output is True if the pair is coreferent and False otherwise. Performance is evaluated using the F1 score in this binary classification setup. Table \ref{tab:gapresults} shows the performance on the GAP test set of the \cite{Lee2017}\footnote{The results that we report for the \cite{Lee2017} system differ slightly from those reported in Table 10 of \cite{gapp} due to a difference in the parser and potentially small differences in the algorithm for converting the system's output to the binary predictions necessary for the GAP scorer.} and \cite{Lee2018} systems as well as the system trained with our adversarial method. The adversarially trained system performs significantly better over the entire dataset in comparison to the previous systems, and the difference is consistent between genders. In particular, we observe that the bias (i.e. ratio of female to male F1 score) is roughly the same ($ 0.93 $) for the \cite{Lee2018} system with and without adversarial training and that this bias is better (i.e. the ratio is closer to $ 1 $) than that exhibited by the \cite{Lee2017} system ($ 0.87 $).

\section{Conclusion}
We show that the performance of the \cite{Lee2018} system decreases when the names of PER and GPE entities are changed in the CoNLL test set so that no names from the training set leak to the test set. We then  retrain the same system using an application of the fast-gradient-sign-method (FGSM) of adversarial training, showing that the retrained system consistently performs better on the original CoNLL test set, the CoNLL test set with No Leakage, and the GAP test set. Our new model is a new state-of-the-art for all these data sets.

\section*{Acknowledgements}
We thank Sihao Chen for providing a gazetteer of first names collected from Wikipedia with scores for their gender likelihood, and the anonymous reviewers for their comments.
This work was supported in part by  contract HR0011-18-2-0052 with the US Defense Advanced Research Projects Agency (DARPA). The views expressed are those of the authors and do not reflect the official policy or position of the Department of Defense or the U.S. Government.

\bibliography{acl2019}

\begin{thebibliography}{22}
\expandafter\ifx\csname natexlab\endcsname\relax\def\natexlab#1{#1}\fi

\bibitem[{Agarwal et~al.(2018)Agarwal, Subramanian, Nenkova, and
  Roth}]{Agarwal18}
Oshin Agarwal, Sanjay Subramanian, Ani Nenkova, and Dan Roth. 2018.
\newblock Named person coreference in english news.
\newblock \emph{arXiv preprint arXiv:1810.11476}.

\bibitem[{Alzantot et~al.(2018)Alzantot, Sharma, Elgohary, Ho, Srivastava, and
  Chang}]{AdvEx}
Moustafa Alzantot, Yash Sharma, Ahmed Elgohary, Bo-Jhang Ho, Mani Srivastava,
  and Kai-Wei Chang. 2018.
\newblock Generating natural language adversarial examples.
\newblock In \emph{Proceedings of the 2018 Conference on Empirical Methods in
  Natural Language Processing}, pages 2890--2896.

\bibitem[{Bekoulis et~al.(2018)Bekoulis, Deleu, Demeester, and
  Develder}]{Bekoulis18}
Giannis Bekoulis, Johannes Deleu, Thomas Demeester, and Chris Develder. 2018.
\newblock Adversarial training for multi-context joint entity and relation
  extraction.
\newblock In \emph{Proceedings of the 2018 Conference on Empirical Methods in
  Natural Language Processing}, pages 2830--2836.

\bibitem[{Chang et~al.(2013)Chang, Samdani, and Roth}]{ChangSaRo13}
Kai-Wei Chang, Rajhans Samdani, and Dan Roth. 2013.
\newblock \href {http://cogcomp.org/papers/ChangSaRo13.pdf} {A constrained
  latent variable model for coreference resolution}.
\newblock In \emph{EMNLP}.

\bibitem[{Ghaddar and Langlais(2016)}]{Wikicoref}
Abbas Ghaddar and Philippe Langlais. 2016.
\newblock Wikicoref: An english coreference-annotated corpus of wikipedia
  articles.
\newblock In \emph{LREC}.

\bibitem[{Goodfellow et~al.(2015)Goodfellow, Shlens, and
  Szegedy}]{Goodfellow14}
Ian Goodfellow, Jonathon Shlens, and Christian Szegedy. 2015.
\newblock \href {http://arxiv.org/abs/1412.6572} {Explaining and harnessing
  adversarial examples}.
\newblock In \emph{International Conference on Learning Representations}.

\bibitem[{Hochreiter and Schmidhuber(1997)}]{lstm}
Sepp Hochreiter and J{\"u}rgen Schmidhuber. 1997.
\newblock Long short-term memory.
\newblock \emph{Neural computation}, 9(8):1735--1780.

\bibitem[{Honnibal and Johnson(2015)}]{spacy}
Matthew Honnibal and Mark Johnson. 2015.
\newblock An improved non-monotonic transition system for dependency parsing.
\newblock In \emph{Proceedings of the 2015 Conference on Empirical Methods in
  Natural Language Processing}, pages 1373--1378.

\bibitem[{Khashabi et~al.(2016)Khashabi, Khot, Sabharwal, Clark, Etzioni, and
  Roth}]{KKSCER16}
Daniel Khashabi, Tushar Khot, Ashish Sabharwal, Peter Clark, Oren Etzioni, and
  Dan Roth. 2016.
\newblock \href {http://cogcomp.org/papers/KKSCER16.pdf} {Question answering
  via integer programming over semi-structured knowledge}.
\newblock In \emph{Proc. of the International Joint Conference on Artificial
  Intelligence (IJCAI)}.

\bibitem[{Lee et~al.(2017)Lee, He, Lewis, and Zettlemoyer}]{Lee2017}
Kenton Lee, Luheng He, Mike Lewis, and Luke Zettlemoyer. 2017.
\newblock End-to-end neural coreference resolution.
\newblock In \emph{Proceedings of the 2017 Conference on Empirical Methods in
  Natural Language Processing}, pages 188--197.

\bibitem[{Lee et~al.(2018)Lee, He, and Zettlemoyer}]{Lee2018}
Kenton Lee, Luheng He, and Luke Zettlemoyer. 2018.
\newblock Higher-order coreference resolution with coarse-to-fine inference.
\newblock In \emph{Proceedings of the 2018 Conference of the North American
  Chapter of the Association for Computational Linguistics: Human Language
  Technologies, Volume 2 (Short Papers)}, pages 687--692.

\bibitem[{McNemar(1947)}]{McNemarTest}
Quinn McNemar. 1947.
\newblock Note on the sampling error of the difference between correlated
  proportions or percentages.
\newblock \emph{Psychometrika}, 12(2):153--157.

\bibitem[{Miyato et~al.(2017)Miyato, Dai, and Goodfellow}]{Miyato16}
Takeru Miyato, Andrew~M. Dai, and Ian Goodfellow. 2017.
\newblock \href {https://arxiv.org/abs/1605.07725} {Adversarial training
  methods for semi-supervised text classification}.
\newblock \emph{ICLR}.

\bibitem[{Moosavi and Strube(2017)}]{Moosavi2017a}
Nafise~Sadat Moosavi and Michael Strube. 2017.
\newblock Lexical features in coreference resolution: To be used with caution.
\newblock In \emph{Proceedings of the 55th Annual Meeting of the Association
  for Computational Linguistics (Volume 2: Short Papers)}, pages 14--19.

\bibitem[{Moosavi and Strube(2018)}]{Moosavi2018b}
Nafise~Sadat Moosavi and Michael Strube. 2018.
\newblock Using linguistic features to improve the generalization capability of
  neural coreference resolvers.
\newblock In \emph{Proceedings of the 2018 Conference on Empirical Methods in
  Natural Language Processing}, pages 193--203.

\bibitem[{Noreen(1989)}]{Noreen89}
Eric~W Noreen. 1989.
\newblock \emph{Computer-intensive methods for testing hypotheses}.
\newblock Wiley New York.

\bibitem[{Pennington et~al.(2014)Pennington, Socher, and Manning}]{glove}
Jeffrey Pennington, Richard Socher, and Christopher Manning. 2014.
\newblock Glove: Global vectors for word representation.
\newblock In \emph{Proceedings of the 2014 conference on empirical methods in
  natural language processing (EMNLP)}, pages 1532--1543.

\bibitem[{Peters et~al.(2018)Peters, Neumann, Iyyer, Gardner, Clark, Lee, and
  Zettlemoyer}]{elmo}
Matthew Peters, Mark Neumann, Mohit Iyyer, Matt Gardner, Christopher Clark,
  Kenton Lee, and Luke Zettlemoyer. 2018.
\newblock Deep contextualized word representations.
\newblock In \emph{Proceedings of the 2018 Conference of the North American
  Chapter of the Association for Computational Linguistics: Human Language
  Technologies, Volume 1 (Long Papers)}, pages 2227--2237.

\bibitem[{Pradhan et~al.(2012)Pradhan, Moschitti, Xue, Uryupina, and
  Zhang}]{Pradhan2012}
Sameer Pradhan, Alessandro Moschitti, Nianwen Xue, Olga Uryupina, and Yuchen
  Zhang. 2012.
\newblock Conll-2012 shared task: Modeling multilingual unrestricted
  coreference in ontonotes.
\newblock In \emph{Joint Conference on EMNLP and CoNLL-Shared Task}, pages
  1--40. Association for Computational Linguistics.

\bibitem[{Srivastava et~al.(2014)Srivastava, Hinton, Krizhevsky, Sutskever, and
  Salakhutdinov}]{Dropout}
Nitish Srivastava, Geoffrey Hinton, Alex Krizhevsky, Ilya Sutskever, and Ruslan
  Salakhutdinov. 2014.
\newblock Dropout: a simple way to prevent neural networks from overfitting.
\newblock \emph{The Journal of Machine Learning Research}, 15(1):1929--1958.

\bibitem[{Webster et~al.(2018)Webster, Recasens, Axelrod, and Baldridge}]{gapp}
Kellie Webster, Marta Recasens, Vera Axelrod, and Jason Baldridge. 2018.
\newblock Mind the gap: A balanced corpus of gendered ambiguou.
\newblock In \emph{Transactions of the ACL}, page to appear.

\bibitem[{Wu et~al.(2017)Wu, Bamman, and Russell}]{Wu17}
Yi~Wu, David Bamman, and Stuart Russell. 2017.
\newblock Adversarial training for relation extraction.
\newblock In \emph{Proceedings of the 2017 Conference on Empirical Methods in
  Natural Language Processing}, pages 1778--1783.

\end{thebibliography}
\bibliographystyle{acl_natbib}

\end{document}